\documentclass[letterpaper, 10 pt, conference]{ieeeconf}  

\IEEEoverridecommandlockouts                              

\usepackage{graphics} 
\usepackage{epsfig} 
\usepackage{times} 
\usepackage{amsmath} 
\usepackage{amssymb}  
\usepackage{graphicx} 
\usepackage{arydshln}
\usepackage[hidelinks]{hyperref}
\hypersetup{
    colorlinks=true,
    linkcolor=black,
    filecolor=black,
    urlcolor=gray,
    breaklinks=true
}
\usepackage{xcolor}
\usepackage{subcaption}
\usepackage{ulem}
\usepackage{booktabs}
\usepackage{makecell}
\usepackage{pifont}

\usepackage{multirow}
\definecolor{mylightgray}{gray}{0.92}
\usepackage{colortbl}

\makeatletter
\renewcommand{\@makefntext}[1]{%
  \parindent 0pt%
  \noindent\hbox{\@makefnmark}~#1%
}
\makeatother

\newcommand\blfootnote[1]{%
  \begingroup
  \renewcommand\thefootnote{}\footnote{#1}%
  \addtocounter{footnote}{-1}%
  \endgroup
}

\title{\LARGE \bf
Improving Generalization of Language-Conditioned \\ Robot Manipulation

}

\author{Chenglin Cui, Chaoran Zhu, Changjae Oh, Andrea Cavallaro%
\thanks{This work was supported in part by the CHIST-ERA program through the project CORSMAL, under the UK EPSRC Grant EP/S031715/1 and in part by the Royal Society Research Grant RGS\textbackslash R2\textbackslash242051. 
This research utilised the Sulis Tier 2 HPC platform under the UK EPSRC Grant EP/T022108/1 and the HPC Midlands+ consortium.
}
\thanks{Chenglin Cui, Chaoran Zhu, and Changjae Oh are with Centre for Intelligent Sensing, Queen Mary University of London, UK.
        {\tt\small \{c.cui,chaoran.zhu,c.oh\}@qmul.ac.uk}. Andrea Cavallaro is with Idiap Research Institute and École Polytechnique Fédérale de Lausanne, Switzerland.
        {\tt\small andrea.cavallaro@epfl.ch}.}}%

\begin{document}

\maketitle
\thispagestyle{empty}
\pagestyle{empty}

\begin{abstract}
The control of robots for manipulation tasks generally relies on visual input. Recent advances in vision-language models (VLMs) enable the use of natural language instructions to condition visual input and control robots in a wider range of environments. However, existing methods require a large amount of data to fine-tune VLMs for operating in unseen environments. In this paper, we present a framework that learns object-arrangement tasks from just a few demonstrations. We propose a two-stage framework that divides object-arrangement tasks into a target localization stage, for picking the object, and a region determination stage for placing the object. We present an instance-level semantic fusion module that aligns the instance-level image crops with the text embedding, enabling the model to identify the target objects defined by the natural language instructions. We validate our method on both simulation and real-world robotic environments. Our method, fine-tuned with a few demonstrations, improves generalization capability and demonstrates zero-shot ability in real-robot manipulation scenarios.
\end{abstract}

\section{Introduction} 

Natural language instructions enable humans to interact more intuitively with robots, enhancing their manipulation capabilities and their ability to generalize across different tasks~\cite{openvla,cliport}. By comprehending and interpreting these instructions, robots can flexibly execute actions based on various commands, even for objects or tasks they have never encountered. To effectively process natural language inputs, robots must also capture the current state of their environments, typically through the visual context. This requirement has led to the integration of Vision-Language Models (VLMs)~\cite{clip, diff} into robotic pipelines, enabling robots to jointly reason over vision and language to guide action execution in real-world scenarios. Two principal approaches have emerged as dominant strategies for integrating VLMs into robotic manipulation tasks, namely instruction-guided object localization~\cite{ma} and optimal scene selection through visual-semantic matching~\cite{ma,dream2real}.

\begin{figure}[t]
  \centering
  \includegraphics[width=0.85\linewidth]{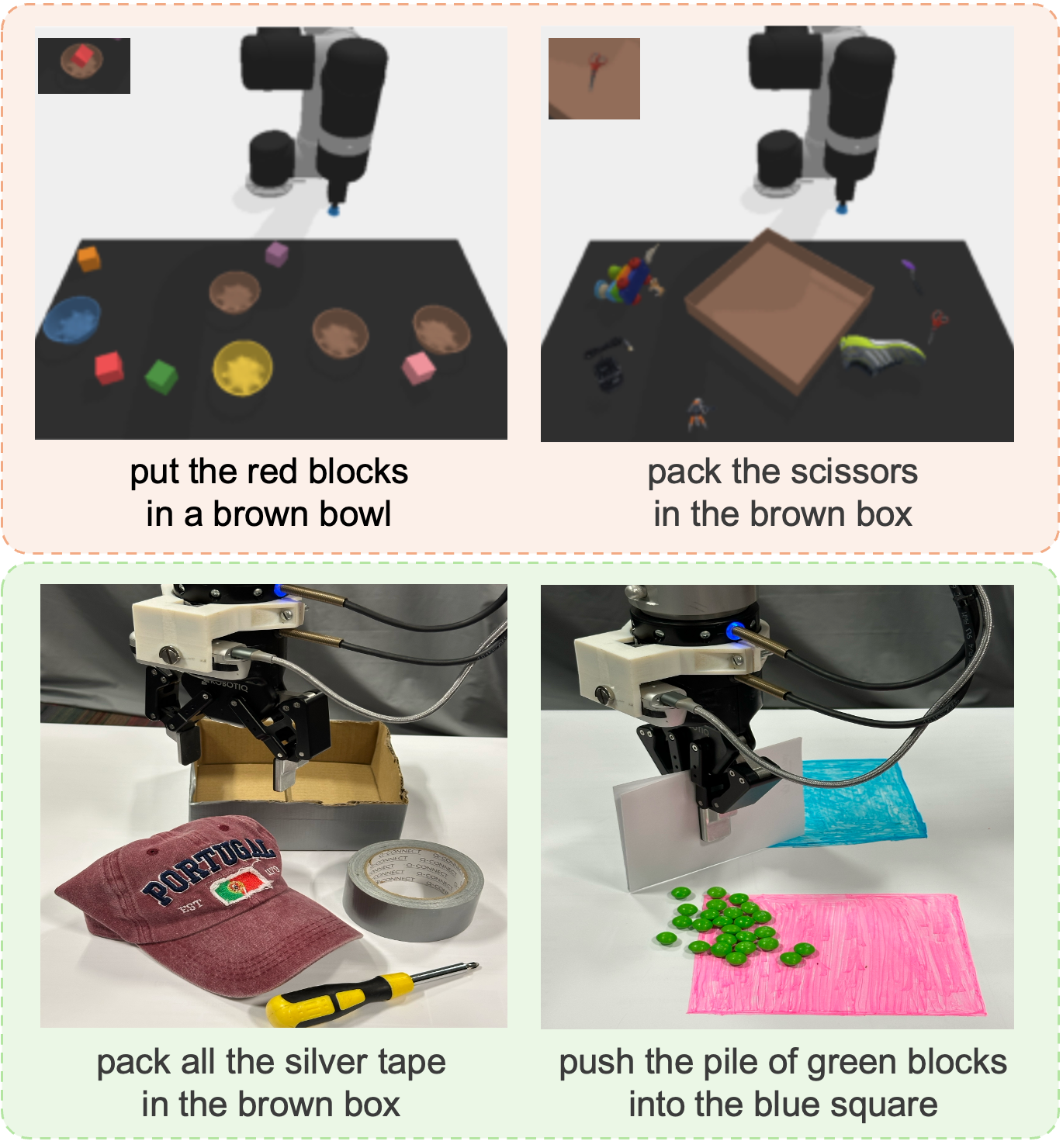}
   \caption{\textbf{Manipulation tasks.} We train our model in a simulated environment and evaluate its performance in both simulation and real-world settings. (Top) Example tasks from the simulation and (bottom) the real-robot environment. }  
  \label{fig: abstract}
\vspace{-9pt}
\end{figure}

To integrate VLMs for specific downstream tasks, existing methods~\cite{cliport,fcclip, conceptfusion} typically freeze the pre-trained VLMs and add trainable downstream decoders. This approach preserves the generalizability of the original VLMs. However, these methods still rely on large amounts of task-specific data to train the decoder from scratch, with the risk of overfitting. Moreover, by keeping the backbone frozen, these models fail to leverage the full potential of VLMs in learning task-relevant representations, ultimately constraining their effectiveness in robotic manipulation. Alternatively, some methods~\cite{cocoop,visualgrounding,clipfit} fine-tune VLMs, add prompts or use knowledge distillation, which can alleviate the domain gap. However, VLMs often emphasize high-level semantic content while neglecting low-level features~\cite{up-vla}, limiting their ability to capture detailed spatial information such as shape and size. Some studies~\cite{lm4lv,igor} highlight that VLMs are particularly weak in low-level vision tasks unless additional training is performed. In our experiments, we observe that VLMs cannot always accurately identify the objects in the environments. For example, while the input command is \textit{pick the gray block}, CLIP~\cite{clip} may localize \textit{the brown block} in the environment, which causes an error in the manipulation task. \blfootnote{The code and video results are available on the website: \url{ https://qm-ipalab.github.io/GeneralPort/}}

To address the aforementioned challenges, we propose a novel framework that enhances VLMs for robotic manipulation through few-shot fine-tuning. We leverage an instance-level semantic fusion strategy to bridge the gap between high-level semantic understanding and low-level spatial perception of objects. This strategy aligns visual and textual features and ensures an accurate perception of object appearance and location. To preserve the generalizability of large pre-trained VLMs, such as CLIP, we fine-tune the model only on a small number of demonstrations (fewer than 20) with a limited set of parameters (fewer than 5M). Experimental results show that our method achieves performance comparable to other methods by training only a small set of parameters and leveraging a few demonstrations.

Our contributions can be summarized as follows:
\begin{itemize}
\item We introduce an instance-level semantic fusion that enhances both high-level semantic understanding and low-level spatial perception for accurate object manipulation.
\item We propose a few-shot fine-tuning strategy for CLIP, enabling task-specific adaptation while preserving its original generalization ability.
\item Our model shows strong zero-shot capability in real-world environments with fewer than 1\% of the trainable parameters of CLIPort~\cite{cliport}.

\end{itemize}

\section{Related Work}

In this section, we discuss cross-modal representations, few-shot learning strategies, and the generalization strategies for robot manipulation tasks.

The integration of \textit{cross-modal representations} through pre-trained VLMs, such as CLIP, has revolutionized robotic manipulation systems~\cite{cliport, taskoriented}, particularly in achieving sub-centimeter localization precision through visual-linguistic alignment.
Several methods~\cite{cliport,programport,gen} use CLIP's visual and text encoders with two-stream architectures to align multimodal features, which requires RGB-D images. 
For task-specific adaptation, GEM~\cite{gen} uses additional training data, including multi-view images, to fine-tune VLMs for robotic policy learning, and VIMA~\cite{vima} uses a visual prompt of the target rather than a text prompt to learn the specific appearance feature of the target. Unlike these methods, our approach does not require additional training data and multi-view visual features.

\textit{Few-shot learning} is a key strategy for enhancing the generalization capability of VLMs in robotic manipulation. Recent approaches focus on efficient fine-tuning techniques, such as parameter-efficient modifications~\cite{clipfit} that adapt large-scale VLMs with minimal adjustments while maintaining their broad generalization capabilities. Proto-CLIP~\cite{proto-clip} further improves few-shot learning by using class prototypes for better task adaptation. DeIL~\cite{deil} introduces the direct-and-inverse CLIP representations to enhance open-world few-shot learning by capturing bidirectional feature mappings. Region Attention Fine-Tuning~\cite{region} utilizes attention and gradient information to automatically identify key entities within support images, generating position prompts. The model is then fine-tuned by minimizing the discrepancy between these position prompts and the attention weights, enabling it to focus on critical regions when processing query images.
Furthermore, LP++~\cite{lp++} demonstrates that simple adaptations can achieve comparable performance with more complex fine-tuning strategies. These approaches highlight the effectiveness of few-shot learning in improving VLMs' adaptability to diverse robotic tasks. Our method also fine-tunes a small subset of the parameters in the pre-trained VLM to maintain its generalization capability. 

\begin{figure*}[t]
  \centering
  \includegraphics[scale=0.58]{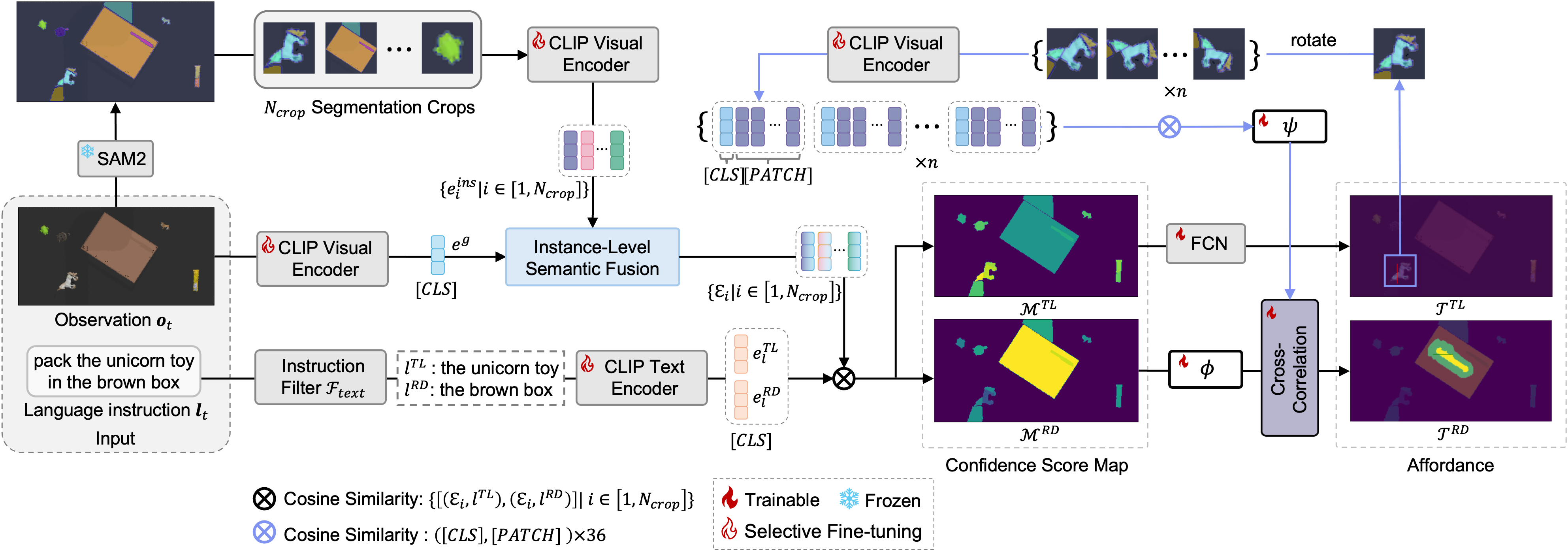}
  \caption{\textbf{Pipeline}. We first input the observation $\mathbf{o}_t$ into the CLIP visual encoder and extract the [CLS] token as a global embedding. Meanwhile, we use SAM2 to segment object masks in $\mathbf{o}_t$ and obtain visual crops of $N_{crop}$ individual objects. These cropped images are then fed into the CLIP visual encoder to generate instance embeddings. To integrate these features, we employ an instance-level semantic fusion module that fuses each instance embedding with the global embedding. Next, we compute a confidence score for each object to assess its likelihood of alignment with the given instruction, which is presented as text embeddings $e^{TL}_l$ and $e^{RD}_l$. In the Target Localization (\textit{TL}) stage, a lightweight fully connected network (FCN) is used to refine the specific manipulation position, ensuring accurate picking. In the Region Determination (\textit{RD}) stage, the crop of the picked object is used to determine the optimal placement position and angle through cross-correlation.}

  \label{fig: pipeline}
\end{figure*}
\textit{Generalization strategies} aim to enhance a robot's performance in unseen environments. CoPa~\cite{copa} introduces a general control policy that incorporates spatial constraints of parts with foundation models. Different loss functions are used for each spatial constraint, and a solver is used to calculate the post-grasp poses based on these constraints. Code as Policies~\cite{codeaspolicy} also leverages a foundation model to generate robot policy code, conditional on natural language commands. To avoid reliance on task-specific training or environment models, ReKep~\cite{rekep} is expressed as Python functions mapping a set of 3D keypoints in the environment to a numerical cost. The method then employs a hierarchical optimization procedure to solve robot actions with a perception-action loop. Moka~\cite{moka} leverages pre-trained VLMs to predict the point-based affordance representations and also converts the motion generation problem into a series of visual question-answering problems that the VLM can address by annotations, such as candidate points, grids, and captions, on RGB images. Our model leverages cross-correlation based on convolution to learn the best placement angle for the target.

\vspace{9pt}
\section{Method}
We consider tabletop object arrangement tasks conditioned on language instructions, observed from the top-down view. Our approach considers the problem of object arrangement as a two-stage task: \textit{target localization (TL)} and \textit{region determination (RD)}. We then introduce an instance-level semantic fusion module that reduces the domain gap between the pre-trained CLIP and the robot manipulation environment. We further present a method to fine-tune our model to a specific task with a few demos, while preserving the generalization capability of CLIP. Finally, we employ a low-level control policy defined over the SE(2) space to perform object grasping based on the predicted pixel location.

\subsection{Object arrangement task} \label{sec: formulation}
Given a visual observation $\mathbf{o}_t$ and language instruction $\mathbf{l}_t$ at timestep \textit{t}, we consider the learning policy $\pi$ as follows:
\begin{equation} 
    \pi(\mathbf{o}_t, \mathbf{l}_t)\rightarrow \mathbf{a}_t = (\mathcal{T}^{\textit{TL}}, \mathcal{T}^{\textit{RD}}) \in \mathcal{A},
\end{equation}
where $\mathcal{T}^{\textit{TL}}$ and $\mathcal{T}^{\textit{RD}}$ are the 2D coordinates for picking and placing, respectively, defined within the affordance space $\mathcal{A}$.
This simple setup contains two essential aspects of the object arrangement problem: (i) there may be a distribution of successful picking poses (from which $\mathcal{T}^{\textit{TL}}$ can be drawn), and (ii) for each successful picking pose, there is a corresponding distribution of successful placing poses (from which $\mathcal{T}^{\textit{RD}}$ can be drawn). We aim to recover these distributions by dividing the object arrangement task into two stages that identify the target object to pick in the \textit{Target Localization stage} and then detect the region to place the object in the \textit{Region Determination stage} conditioned on the outcome of the first stage.

In the \textit{Target Localization stage}, we design a policy module $\mathcal{Q}^{\textit{TL}}$ using Fully Convolutional Networks (FCNs) and predict the picking pose distribution $\mathcal{T}^{\textit{TL}}$:
\begin{equation}
    \mathcal{T}^{\textit{TL}} = \arg\max_{(u,v) \in \mathbf{o}_t} \mathcal{Q}^{\textit{TL}}((u,v)| \mathbf{o}_t, \mathbf{l}_t).
\end{equation}
The visual observation $\mathbf{o}_t$ is a top-down view image, and hence each pixel location $(u,v)$ can be mapped to a 3D picking location using the known camera calibration. 

In the \textit{Region Determination stage}, 
we first extract $c \times c$ size image crop of picked target from $\mathbf{o}_t$ centered at $\mathcal{T}^{\textit{TL}}$. 
As the given observation $\mathbf{o}_t$ is an orthographic heightmap, the visual presentation of rigid objects remains constant, so that we retrieve the angle of object placement conditioned on the picked object. As illustrated in the upper-right section of Fig.~\ref{fig: pipeline}, we leverage the structure of a rigid object from pixel mask to the best region in the orthographic heightmap during the region determination stage. We rotate the selected object image by $\tau_j$ degrees to produce corresponding cropped images. Here, $\{\tau_j \mid 1 \leq j \leq n\}$ represents a set of discrete rotation angles, with $n = 36$.
We then use two FCNs $[\psi,\phi]$ to obtain the embeddings of $\mathbf{o}_t$ and $n$ image crops, respectively. The policy module  $\mathcal{Q}^{\textit{RD}}$ computes the cross-correlation between the embeddings from $[\psi,\phi]$:
\begin{equation}
    \mathcal{Q} ^{\textit{RD}}(\tau|\mathbf{o}_t, \mathbf{l}_t, \mathcal{T}^{\textit{TL}}) = \psi(\mathbf{o}_t[\mathcal{T}^{\textit{TL}}]) * \phi(\mathbf{o}_t)[\tau], 
\end{equation}
where $\tau$ is the angle rotation of the crop, $*$ represents the  cross-correlation operation, which is a modified convolution operation that uses $n$ rotated crop features as convolution kernels to perform convolution operations on the feature maps of $\mathcal{M}^{RD}$. Then we can get the pose by maximizing the policy:
\begin{equation}
    \mathcal{T}^{\textit{RD}} = \arg\max_{\tau_j =10 j} \mathcal{Q} ^{\textit{RD}}(\tau_j| \mathbf{o}_t, \mathbf{l}_t, \mathcal{T}^{\textit{TL}}),
\end{equation}
To divide the text instruction into a target object description $l^{\textit{TL}}$ and a placement region description $l^{\textit{RD}}$, we use a manually designed text filtering mechanism $\mathcal{F}_{text}$. For example, the input instruction \textit{put the brown blocks in a cyan bowl} could be separated into \textit{a photo of the brown blocks} for the \textit{TL} stage and \textit{a photo of a cyan bowl} for the \textit{RD} stage after the filtering. 

\subsection{Instance-level semantic fusion} \label{sec: instance}
We use a pre-trained CLIP model to identify the object within the visual observation that is most relevant to the current task. Specifically, the pre-trained CLIP takes the visual embeddings of $\mathbf{o}_t$ and the segmented object $b_i$ ($ i \in[1,N_{crop}]$) as input and generates an instance-level semantic embedding $\mathcal{E}$. 

Before instance-level fusion, we leverage SAM2~\cite{sam2} to generate the mask and bounding box of every single object in the observation and extract its visual embedding $e_{i}^{ins}$ along with the global embedding $e^{g}$ of the entire observation, which provides context information for the overall scene.

Using $\omega_i \in [0,1]$ as the weight to fuse the two embeddings, we calculate the instance-level embedding based on $e_{i}^{ins}$ and $e^{g}$ as follows:
\begin{equation} 
    \mathcal{E}_i = \omega_i e^g + (1-\omega_i)e^{ins}_i.
\end{equation}
To enable $\omega_i$ to better represent the relationship between each object and the whole observation, we compute two key similarities $\zeta$ and $\eta$. We first compute the cosine similarity $\zeta$ between each instance embedding and the global embedding:
\begin{equation}
    \zeta_i = \left \langle e_{i}^{ins}, e^{g} \right \rangle= \frac{e_{i}^{ins} \cdot e^{g} }{|e_{i}^{ins}|\ |e^{g} |},
\end{equation}
and a matrix of cosine similarities between all pairs of instance embeddings as $\eta_{ij} = \left \langle e_{i}^{ins}, e_{j}^{ins} \right \rangle $. We then compute the average similarity $\eta_i$ with all other object embeddings for each object embedding, which could capture the difference between them, thus accounting for the uniqueness of the instance for $e_{i}^{ins}$ in the image:
\begin{equation} 
    \eta_i = \frac{1}{N_{crop}} \sum^{N_{crop}}_{j=1,j\neq i} \eta_{ij},
\end{equation}
where $N_{crop}$ is the number of all segmentation crops.
We then combine $\zeta_i$ and $\eta_i$ to compute $\omega_i \in [0,1]$ as:
\begin{equation} 
    \omega_i = \frac{exp\left( \frac{\zeta_i + \eta_i}{\tau} \right)}{\sum^R_{i=1}exp\left( \frac{\zeta_i + \eta_i}{\tau} \right)} ,
\end{equation}
%
where \(\tau\) is a temperature parameter that adjusts the smoothness of the softmax distribution.
We then calculate the cosine similarity between fused embeddings $\mathcal{E}_i$ from the instance-level semantic fusion and the text embeddings $(e^{\textit{TL}}_l, e^{\textit{RD}}_l)$ respectively from the CLIP text encoder to obtain the confidence score $s_i$. The value of each pixel of the $i$-th object's mask is set to $s_i$ to construct the confidence score map $(\mathcal{M}^{\textit{TL}}, \mathcal{M}^{\textit{RD}})$ as shown in Fig.~\ref{fig: pipeline}.

\subsection{Few-shot fine-tuning} \label{sec: finetune}

CLIP consists of a visual encoder and a text encoder~\cite{clip}. Without introducing any external parameters, we only fine-tune the bias terms of projection linear layers in the Feed Forward Networks (FFNs) of the text encoder and update the LayerNorm in the visual encoder. For the text encoder, instead of fine-tuning all bias terms, we fine-tune only the bias terms of the projection linear layers in the FFNs of the text encoder, which will reduce the number of training parameters. Moreover, only fine-tuning the projection linear layers' bias terms could achieve better performance compared with fine-tuning all the bias terms~\cite {clipfit}.

For the visual encoder, we fine-tune LayerNorm instead of fine-tuning the bias terms. In LayerNorm, the two learnable parameters gain and bias are applied for an affine transformation on normalized input vectors for re-centering and re-scaling, which is crucial for adapting to new data distributions and enhancing the expressive power. Different data distributions should produce different gains and biases in LayerNorm for distribution reshaping during the training process. So, if shifted gains and biases in LayerNorm are applied during inference, it may lead to a sub-optimal solution. Therefore, we only fine-tune LayerNorm in the image encoder.

\begin{figure}[t]
    \centering
    \begin{subfigure}{0.23\textwidth} 
        \centering
        \includegraphics[width=\linewidth]{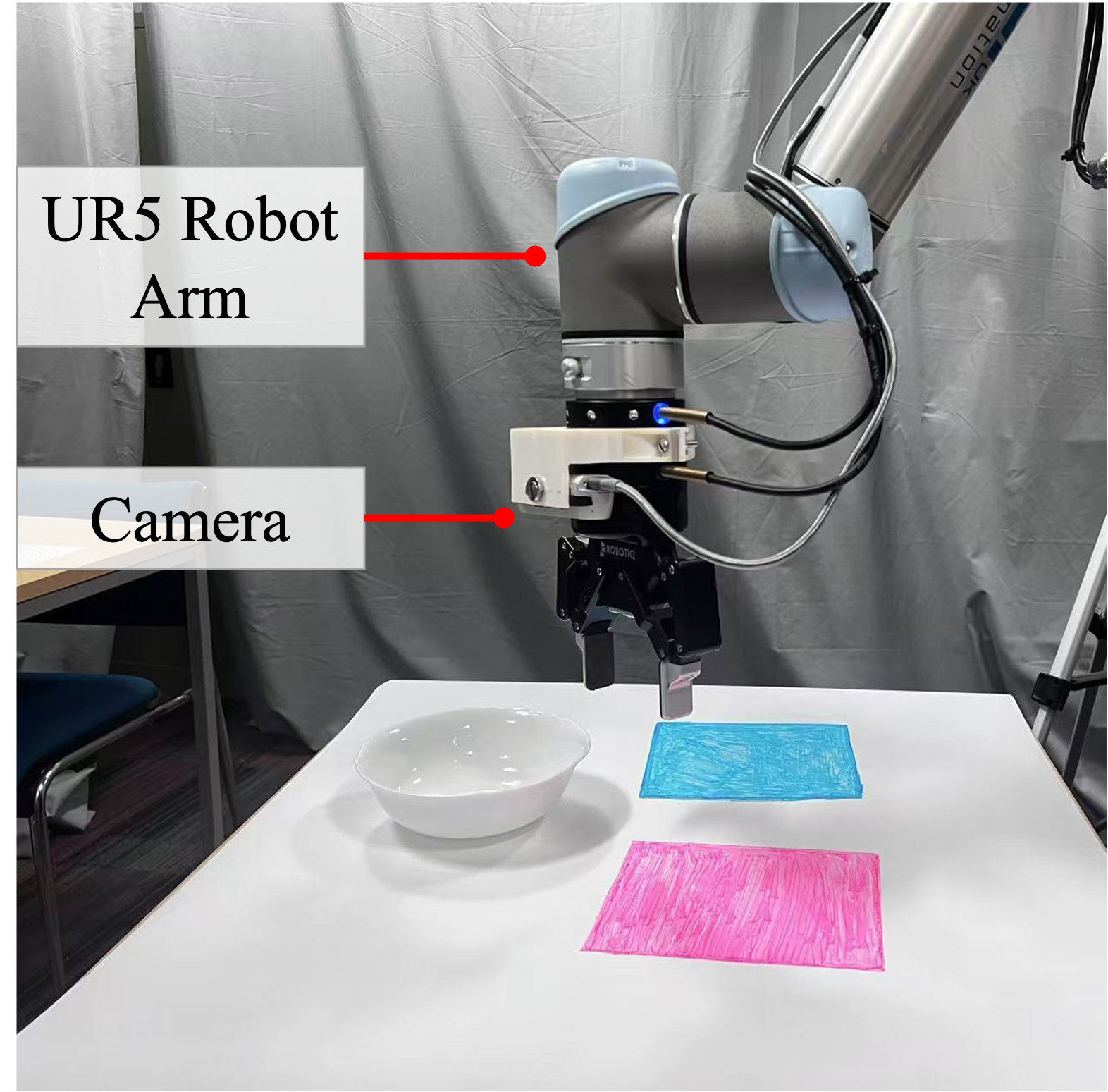}
        \caption{Real-robot setup}
    \end{subfigure}
    \hfill
    \begin{subfigure}{0.23\textwidth}
        \centering
        \includegraphics[width=\linewidth]{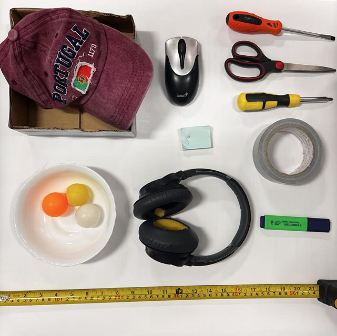}
        \caption{Objects for testing}
    \end{subfigure}
\caption{\textbf{Real-robot experiment setup.} All experiments were conducted using a UR5 robot arm equipped with a Robotiq 2F-85 2-finger gripper and an Intel Realsense D435 camera mounted on the robot arm. Our model relies solely on RGB images, while CLIPort requires RGB-D images. Both types of images are captured from a top-down view. The robotic arm first moves to the initial top-down position to capture the input image and then proceeds with the manipulation.}
\label{fig:setup}

\end{figure}

\section{Validation}
\subsection{Setup}\label{sec:setup}
\textbf{Simulation environment}.
We evaluate our method in a simulation environment for fair and reproducible comparisons. We use the benchmark proposed by CLIPort~\cite{cliport} built on PyBullet Gym~\cite{gym}.
From this benchmark, we select nine diverse manipulation tasks in the benchmark, covering both seen and unseen scenarios, and measure the success rates. We utilize SAM2~\cite{sam2} to extract both object masks and their bounding boxes from the input observation. To encode the segmented observations, we employ the pre-trained CLIP with the ViT-H/14 backbone~\cite{clip} for visual feature extraction, while the ViT-B/16 model is used to encode the crop of a picked object.
We supervise our model using cross-entropy loss during both the TL stage and the RD stage. The ground truth in the Benchmark is a one-hot encoded position vector over all pixels in the observation, while the predicted vector is a distribution vector over these pixels. The overall loss function of our model is a weighted sum of both losses from the TL stage and the RD stage.

\begin{table*}[!t]\large
\centering
\vspace{9pt}
\caption{\textbf{Performance on the CLIPort benchmark~\cite{cliport}}. We measure the success rates of object arrangement tasks across nine different environments where objects or their colors were unseen. The models are fine-tuned with 1, 10 and 20 demonstrations, and each task is tested with 50 episodes.}
\renewcommand{\arraystretch}{1.1} 
\aboverulesep=0ex
\belowrulesep=0ex
\label{tab: simulation results}
\resizebox{\textwidth}{!}{
\begin{tabular}{lccccccccccccccc}
\hline
 & \multicolumn{3}{c}{} & \multicolumn{3}{c}{\begin{tabular}[c]{@{}c@{}}packing-box-pairs\\ \textcolor{red}{unseen-colors}\end{tabular}} & \multicolumn{3}{c}{\begin{tabular}[c]{@{}c@{}}packing-\textcolor{blue}{unseen-google}\\ \textcolor{blue}{objects}-seq\end{tabular}} & \multicolumn{3}{c}{\begin{tabular}[c]{@{}c@{}}packing-\textcolor{blue}{unseen-google}\\ \textcolor{blue}{objects}-group\end{tabular}}  
 & \multicolumn{3}{c}{packing-\textcolor{blue}{unseen-shapes}} \\
   \cmidrule(lr){5-7}
   \cmidrule(lr){8-10}
   \cmidrule(lr){11-13}   
   \cmidrule(r){14-16}
   \textbf{Method} & 
   \multicolumn{3}{c}{\textbf{Input}}&
  {1} & 10 &20 & 
  {1} & 10 &20 & 
  {1} & 10 &20 & 
  {1} & 10 &20   \\ 
  \cline{1-16}
TransporterNet &  \multicolumn{3}{c}{RGB-D} & \cellcolor{mylightgray}34.6 & 34.6 & 40.4 & \cellcolor{mylightgray}4.9 & 31.7 & 46.8 & \cellcolor{mylightgray}20.2 & 46.0 & 56.5 & \cellcolor{mylightgray}4.0 & 24.0 & 26.0 \\
CLIPort &  \multicolumn{3}{c}{RGB-D}  & \cellcolor{mylightgray}1.9 & 44.3 & \textbf{64.2} & \cellcolor{mylightgray}3.6 & 25.2 & 45.9 & \cellcolor{mylightgray}37.1 & 46.6 & 50.3 & \cellcolor{mylightgray}16.0 & 29.0 & 24.0 \\
Ours &  \multicolumn{3}{c}{RGB}  &\cellcolor{mylightgray}\textbf{{36.9}} & \textbf{45.6} & 38.3 & \cellcolor{mylightgray}\textbf{{48.9}} & \textbf{45.2} & \textbf{50.2} & \cellcolor{mylightgray}\textbf{{45.4}} & \textbf{54.6} & \textbf{57.3} & \cellcolor{mylightgray}\textbf{{46.0}} & \textbf{50.0} & \textbf{46.0} \\ 
\hline
   & \multicolumn{3}{c}{\begin{tabular}[c]{@{}c@{}}put-block-in-bowls\\ \textcolor{red}{unseen-colors}\end{tabular}} & \multicolumn{3}{c}{\begin{tabular}[c]{@{}c@{}}towers-of-hanoi-seq\\ \textcolor{red}{unseen-colors}\end{tabular}} & \multicolumn{3}{c}{\begin{tabular}[c]{@{}c@{}}stack-block-pyramid\\ seq-\textcolor{red}{unseen-colors}\end{tabular}} & \multicolumn{3}{c}{\begin{tabular}[c]{@{}c@{}}assembling-kits-seq\\ \textcolor{red}{unseen-colors}\end{tabular}} &\multicolumn{3}{c}{\begin{tabular}[c]{@{}c@{}}separating-piles\\ \textcolor{red}{unseen-colors}\end{tabular}} \\ 
   \cmidrule(lr){2-4} 
   \cmidrule(lr){5-7}
   \cmidrule(lr){8-10}
   \cmidrule(lr){11-13}
   \cmidrule(lr){14-16}
 \textbf{Method}  & 
  {1} & 10 &20 & 
  {1} & 10 &20 & 
  {1} & 10 &20 & 
  {1} & 10 &20 & 
  {1} & 10 &20 \\ \hline
TransporterNet &  \cellcolor{mylightgray}4.0 & 12.0 & 17.0 & \cellcolor{mylightgray}2.8 & 43.4 & \textbf{48.3} & \cellcolor{mylightgray}0.3 & 18.0 & \textbf{26.0} & \cellcolor{mylightgray}0.0 & \textbf{20.0} & \textbf{24.8} & \cellcolor{mylightgray}12.4 & 41.5 & 19.9 \\
CLIPort &  \cellcolor{mylightgray}6.0 & 50.8 & 41.7 & \cellcolor{mylightgray}10.3 & \textbf{57.6} & 47.3 & \cellcolor{mylightgray}0.0 & \textbf{21.8} & 19.2 & \cellcolor{mylightgray}0.0 & 16.6 & 20.6 & \cellcolor{mylightgray}11.2 & 43.8 & 49.6 \\
Ours &  \cellcolor{mylightgray}\textbf{{82.0}} & \textbf{83.6} & \textbf{92.6} & \cellcolor{mylightgray} \textbf{{14.3}}& 11.5 & 11.4 & \cellcolor{mylightgray}\textbf{{2.6}} & 7.3 & 12.0 & \cellcolor{mylightgray}\textbf{{4.8}} & 12.8 & 13.6  & \cellcolor{mylightgray}\textbf{{47.5}} & \textbf{48.2} & \textbf{51.2}\\ 
\hline

\end{tabular}
}
\end{table*}

\begin{figure*}[t]
  \centering
  \includegraphics[scale=0.4]{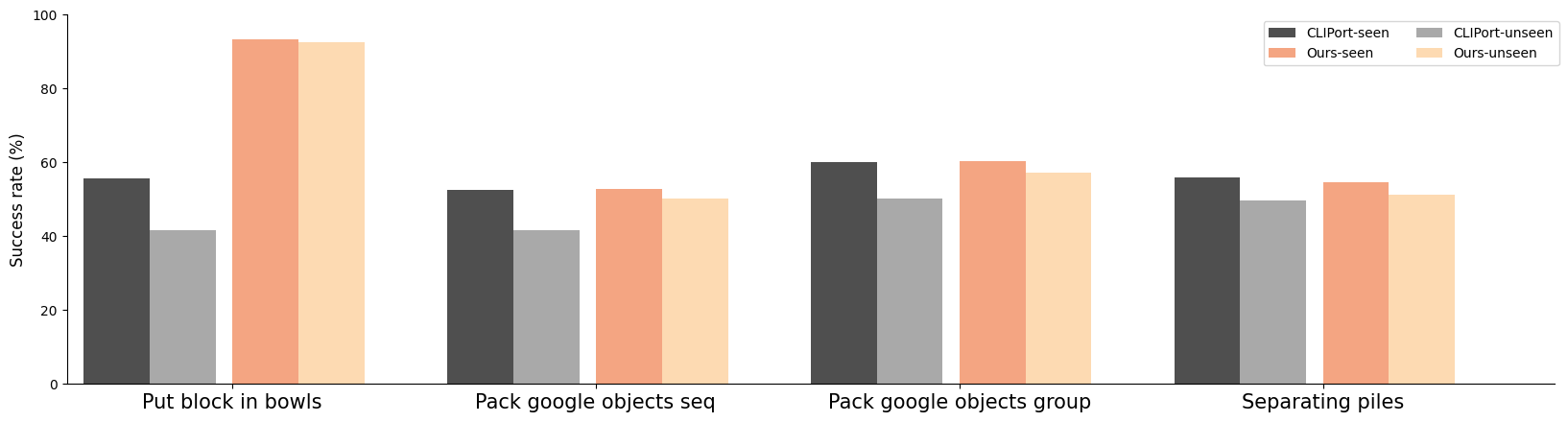}
  \caption{\textbf{Results on seen and unseen tasks in simulation.} We present the results of four different tasks, each trained for 20 episodes, categorized into seen and unseen groups.}
  \label{fig:gap}
\end{figure*}

\textbf{Real-robot environment}. 
In Fig.~\ref{fig:setup}, we show our real-robot environment. The real-robot setup includes a UR5 robot arm with a Robotiq 2F-85 2-finger gripper and an Intel Realsense D435 camera serving as a hand-eye camera. We validate our method on three manipulation tasks named: \textit{put-balls-in-bowl}, \textit{pack-objects-in-box}, and \textit{separating-piles}. Each task consists of five different layouts to ensure the variation in the object arrangements. To improve inference efficiency, we replace the original CLIP model with ViT-B/16 and adopt a lightweight variant of SAM2 for segmentation. For each task, we set the maximum number of step to 5 for the whole process to allow multiple attempts to complete the task. We calculate the success rate as an evaluation metric.  


\textbf{Baseline models}. 
We compare our method with two established baselines, Transporter~\cite{transporternet} and CLIPort~\cite{cliport}. We do not compare our method with ProgramPort~\cite{programport} and GEM~\cite{gen}, as these methods extend the CLIPort benchmark by using additional training data, whereas our method does not require any extra training data.

\subsection{Analysis}
\textbf{Simulation results}.
In the simulation environment, both our model and the baselines are trained using 1, 10, and 20 demonstrations. Table~\ref{tab: simulation results} shows the results on unseen environments where the objects or their color do not appear in training. The results show that our model consistently outperforms the baselines across most tasks, often by a substantial margin. Moreover, our model trained on a single demonstration outperforms existing models across all tasks, which demonstrates superior sample efficiency. In Fig.~\ref{fig:gap}, we analyze the gap between the seen and unseen environments in four tasks. The performance gap of CLIPort between the seen and unseen environments is larger than that of our method, indicating that our method has better generalization capability. In addition, our model has only 4.6M trainable parameters, significantly less than CLIPort’s trainable parameters (593M) and TransporterNet's (29.5M).

\begin{figure*}[t]
  \centering
  \footnotesize
  \setlength{\tabcolsep}{0.8pt}
  \newcommand{\sz}{0.115}
  \begin{tabular}{lcccc}
    & \makecell{``Push the pile of yellow blocks \\ into the blue square"} & \makecell{``Pick the silver tape \\ in the brown box "}& \makecell{``Pick the yellow ball \\ in the white bowl"} & \makecell{``Pick the baseball hat \\ in the brown box"} \\
    \makecell{\rotatebox{90}{Input image}}&
    \makecell{\includegraphics[height=\sz\linewidth]{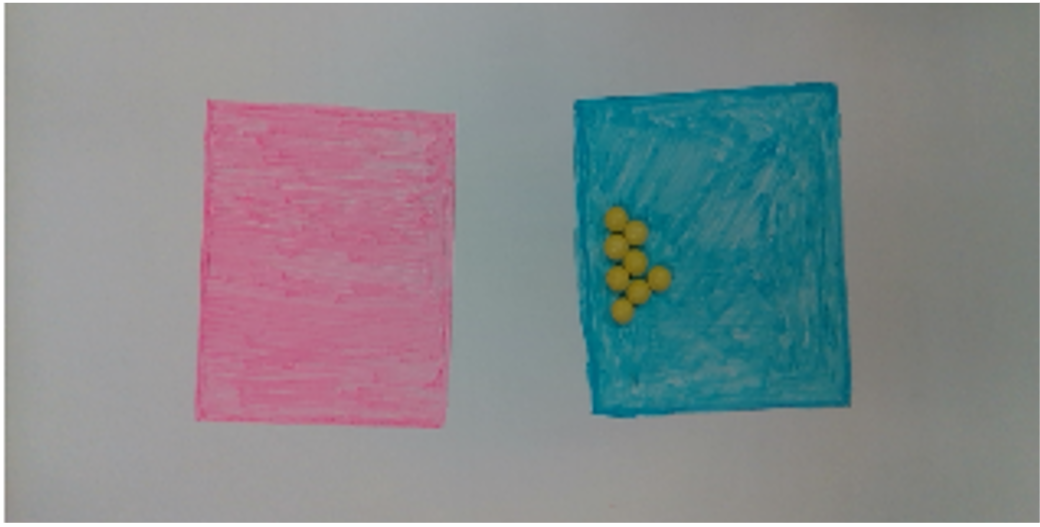}} &
    \makecell{\includegraphics[height=\sz\linewidth]{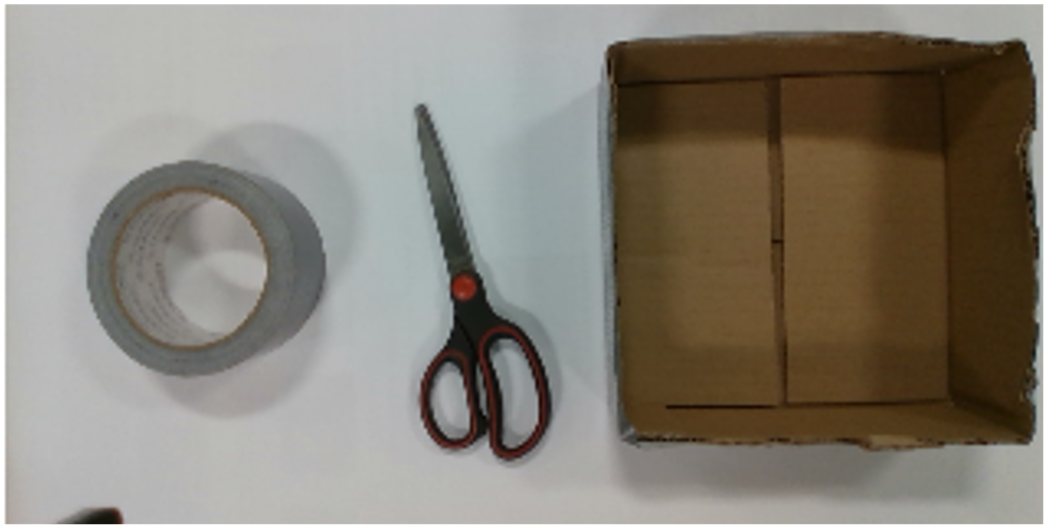}} &
    \makecell{\includegraphics[height=\sz\linewidth]{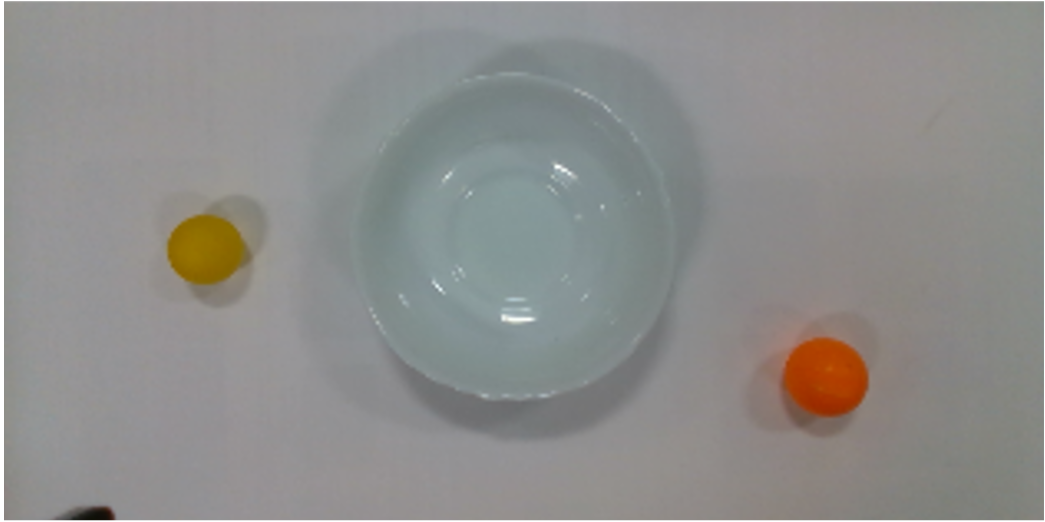}} &
    \makecell{\includegraphics[height=\sz\linewidth]{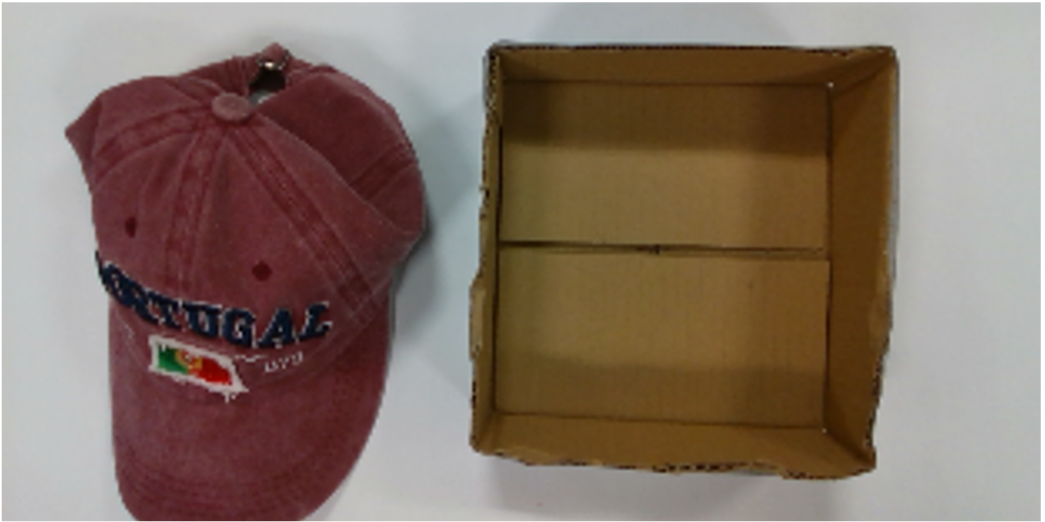}}\\

    \makecell{\rotatebox{90}{Pick / Push from}}&
    \makecell{\includegraphics[height=\sz\linewidth]{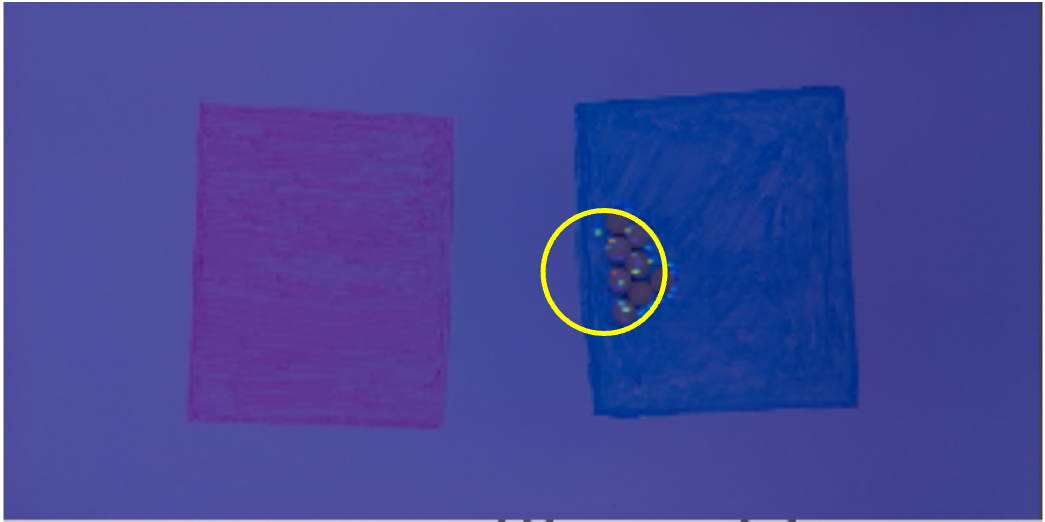}} &
    \makecell{\includegraphics[height=\sz\linewidth]{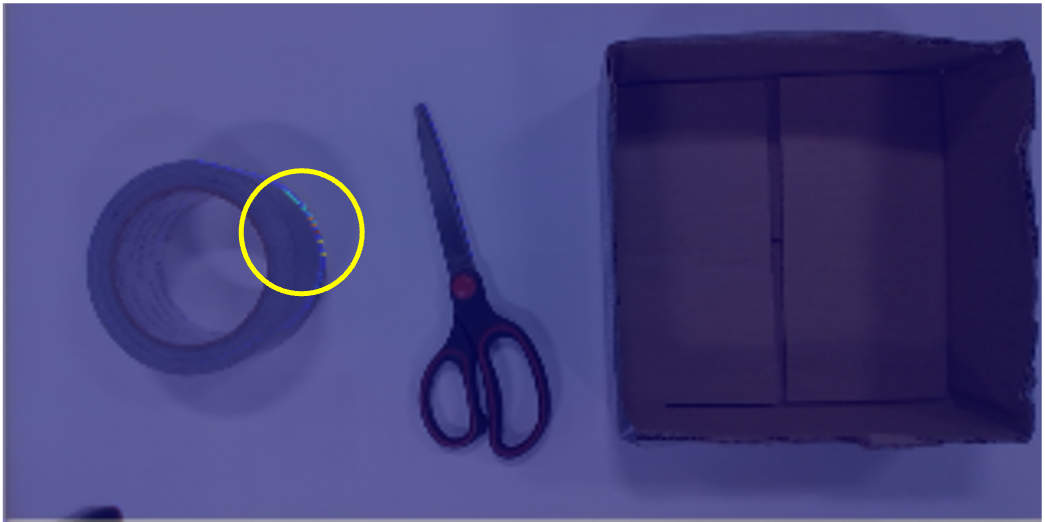}} &
    \makecell{\includegraphics[height=\sz\linewidth]{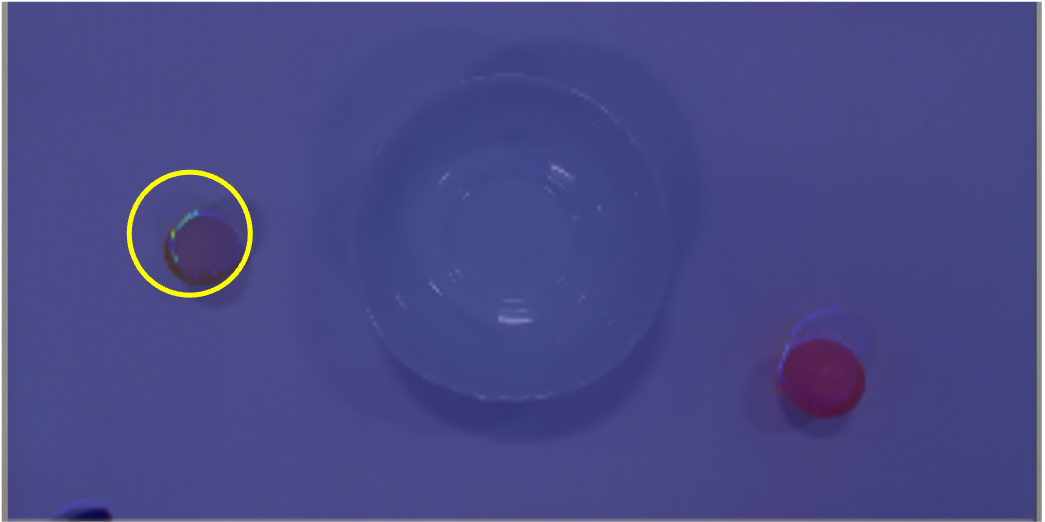}} &
    \makecell{\includegraphics[height=\sz\linewidth]{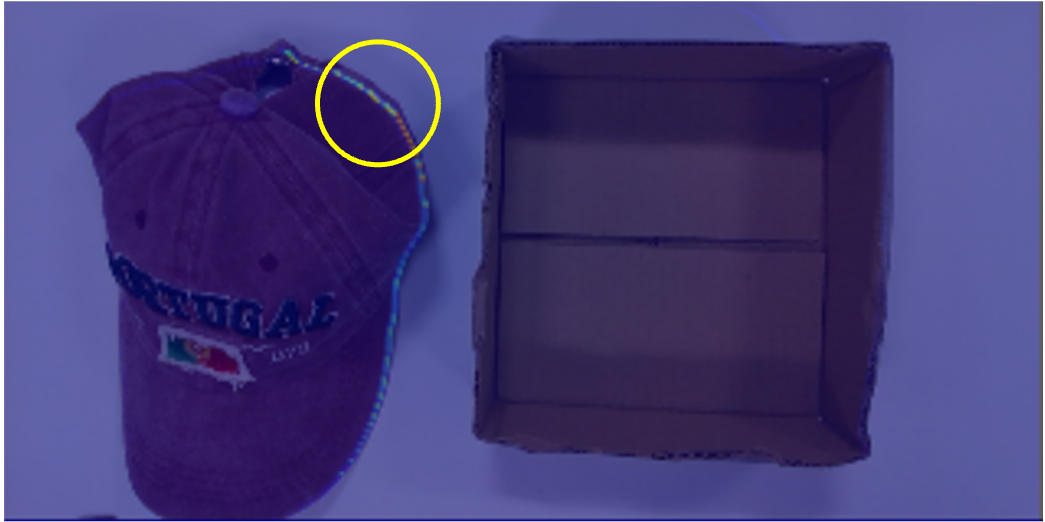}}\\
    
    \makecell{\rotatebox{90}{Place / Push to}}&
    \makecell{\includegraphics[height=\sz\linewidth]{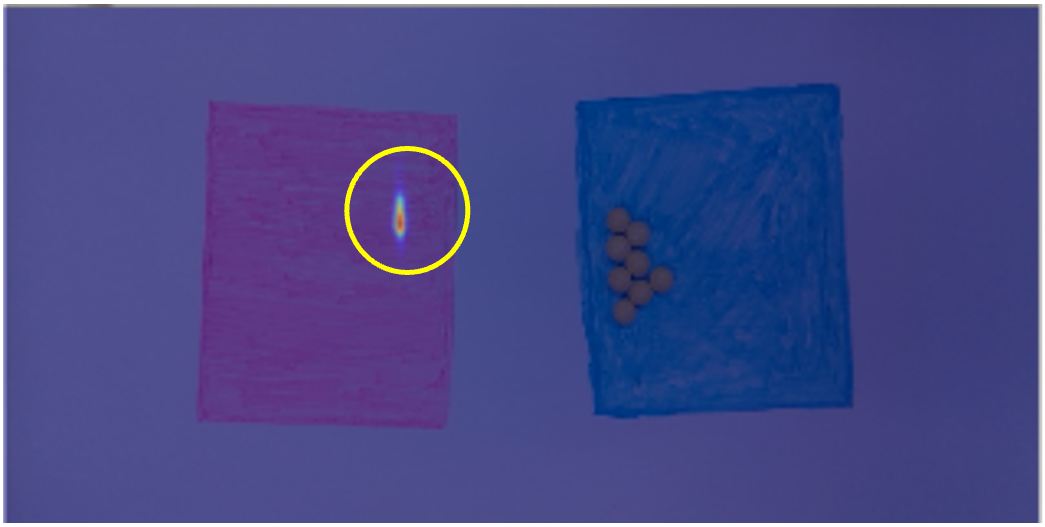}} &
    \makecell{\includegraphics[height=\sz\linewidth]{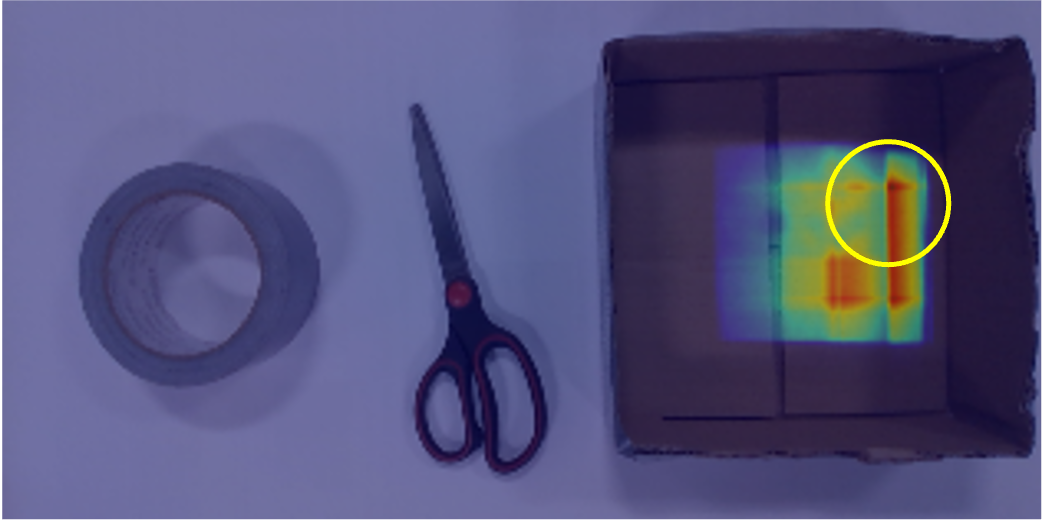}} &
    \makecell{\includegraphics[height=\sz\linewidth]{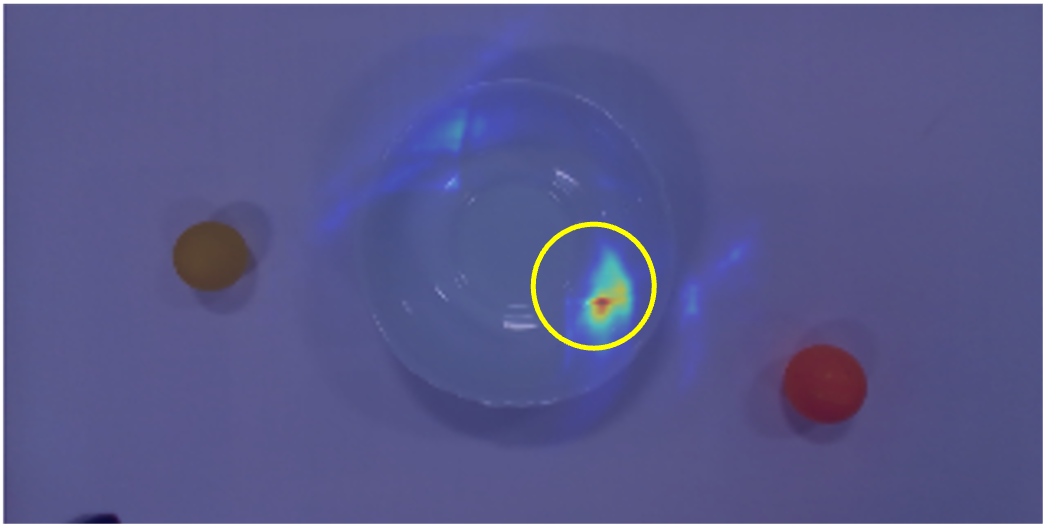}} &
    \makecell{\includegraphics[height=\sz\linewidth]{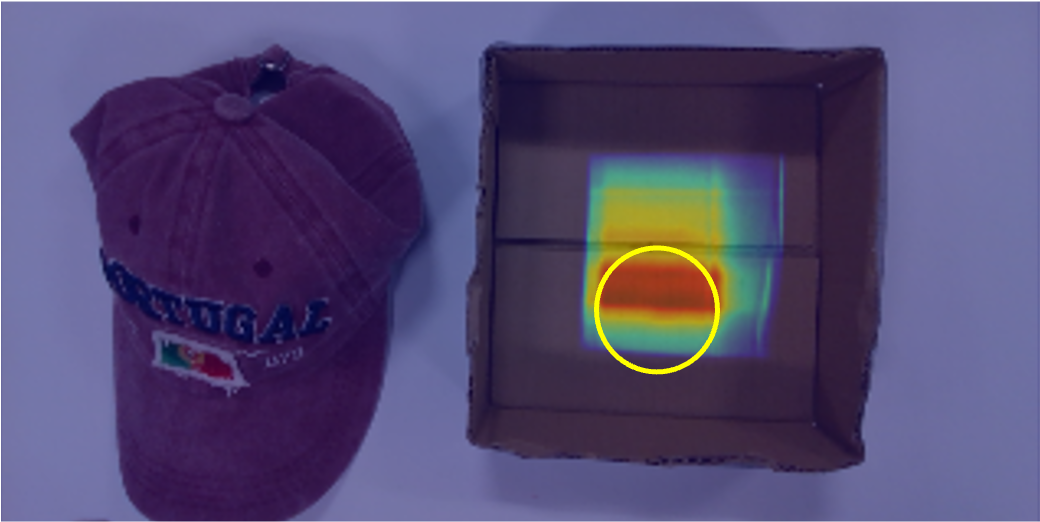}}\\
  \end{tabular} 
  \caption{\textbf{Qualitative examples from the real-robot environment.} We visualize the affordance maps of ``Pick/Push from ...'' and the ``Place/Push to ...'', which highlight the locations of the relevant target objects (in yellow circle). These affordance maps represent a probability distribution over potential manipulation places, indicating not only the target objects but also the most suitable areas for robotic interaction. The corresponding rotation angles are not depicted in the images.}
  \label{fig:Affordance}
  
\end{figure*}

\begin{table}[t]\large
\centering
\caption{\textbf{Real-robot experiment.} Success rates of three arrangement tasks in real-robot environments. Our method exhibits strong generalization even in a zero-shot setting, while CLIPort struggles to perform well, even after being trained on a set of real-world data.}
\label{tab:real_robot}
\resizebox{\columnwidth}{!}{%
\begin{tabular}{@{}clccc@{}}
\toprule
Model   & zero-shot & put-balls-in-bowl & pack-object-in-box & separating-piles \\ \midrule
CLIPort & {\makecell{\ding{55}}} & 0.0               & 50.0               & 36.0             \\
Ours    & {\makecell{\ding{51}}} & 74.0              & 86.1               & 40.0             \\ \bottomrule
\end{tabular}%
}

\end{table}

\textbf{Real-robot results.}\label{sec: real-robot}
Table~\ref{tab:real_robot} shows the success rates of three object arrangement tasks in the real-robot environment. Our model is tested in a zero-shot setting, where the model is only trained on the simulation environment and tested directly on the real-robot environment. Our method achieves 74.0\%, 86.1\%, and 40.0\% on the three tasks since our model leverages SAM2 and CLIP, which are trained on large-scale real-world data. In Fig.~\ref{fig:Affordance}, the results show that our model can handle the real-robot environment.
In contrast, CLIPort failed to perform the tasks without fine-tuning on real-robot data. We hence report the results of CLIPort after fine-tuning with 30 demonstrations, which still underperform our method in the zero-shot setting.

\begin{figure}[t]
    \centering
  \includegraphics[scale=0.75]{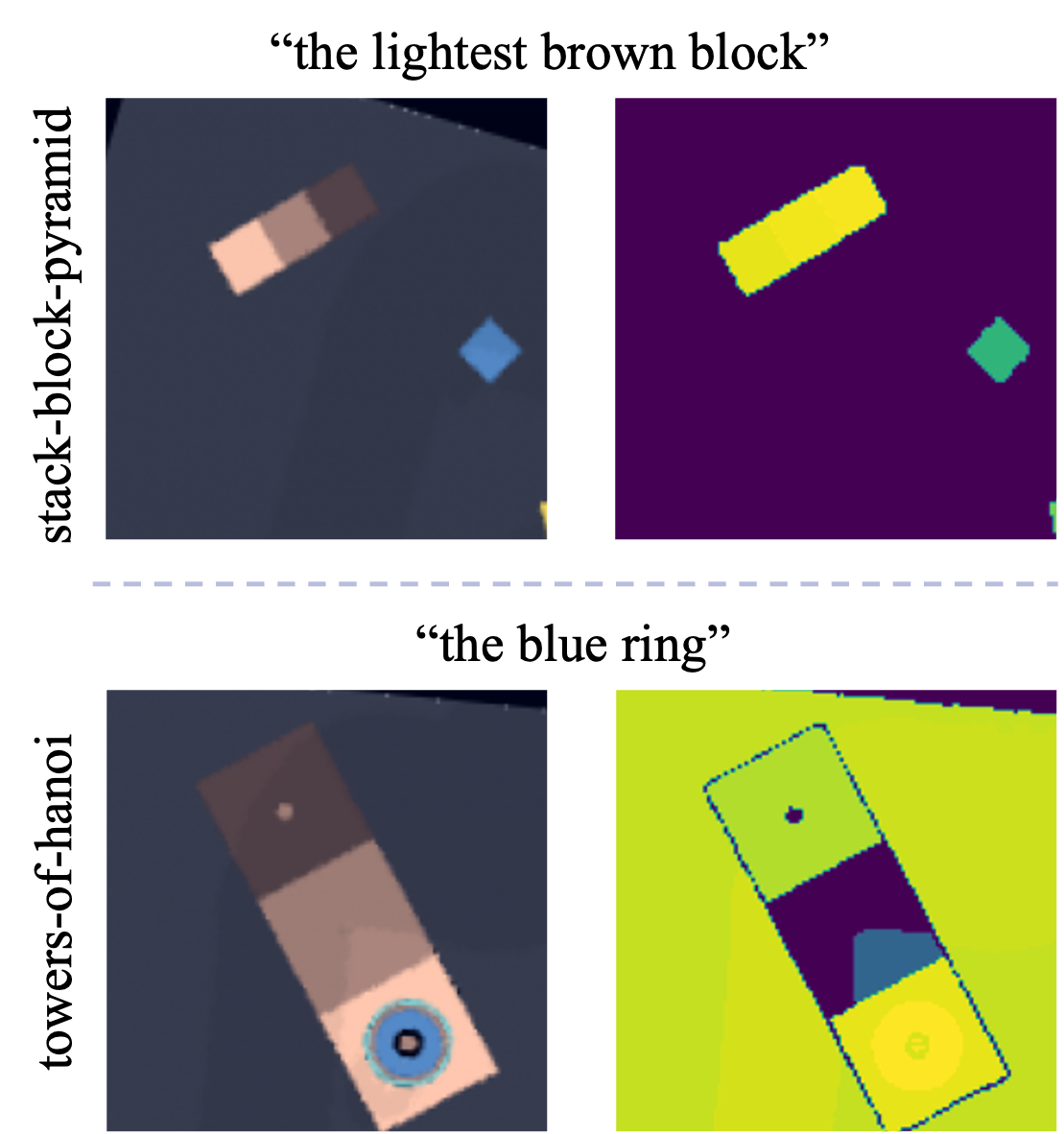}
  \caption{\textbf{Sample failure cases}.Due to the inherent limitations of CLIP, similar colors can interfere with the alignment between visual and textual representations. As the bottom row shows, it is difficult to accurately localize the boundary of the target because of the limitations of SAM2, while the target picking pixel of affordance map in training data is very close to the boundary.} 
  \label{fig: failed cases}  
 \vspace{-9pt}

\end{figure}

\textbf{Failure cases.} 
Due to the limitations of the CLIP model, our model suffers from discriminating against similar colors. For instance, in the \textit{stack-block-pyramid} task in Fig.~\ref{fig: failed cases}, the result shows that the model cannot output the highest confidence score on the region of \textit{the lightest brown block}. This result indicates that the model struggles to accurately localize the object most relevant to the input instruction and provides incorrect information to the downstream FCN. Furthermore, the performance of the segmentation model also affects the results of our method. The \textit{towers-of-hanoi} task in Fig.~\ref{fig: failed cases} shows that SAM2 cannot accurately segment \textit{the blue ring} due to multiple stacked rings. This segmentation error may result in further errors in predicting the final picking point. As our current method relies on SAM2 for segmentation, fine-tuning a segmentation model can address the problem by generating fine-grained masks of objects.

\section{Conclusion}
We addressed the challenge of improving generalization in robot manipulation tasks, specifically unseen colors and unseen objects. Our two-stage framework divides an object arrangement task into the target localization stage and the region determination stage, using separate networks for picking and placing objects. The proposed instance-level semantic fusion module enables the model to align the segmented objects with the language instructions to bridge the domain gap. To adapt to unseen environments, our method is trained with a few demonstrations and selectively fine-tunes a small portion of CLIP parameters to preserve generalization. The simulation results demonstrate that our method outperforms existing ones, especially when the demonstration becomes sparse. Our method, trained in the simulation environment, further showed the zero-shot capability in a real-robot environment. In future work, we will reduce reliance on the pre-trained segmentation model by incorporating more robust perception mechanisms and extend our approach to execute full 6-DoF robot actions to perform more complex manipulations in real-world environments. Replacing manual text parsing with large language models can streamline the language understanding process and improve robustness.

\addtolength{\textheight}{-10cm}   




{
\bibliographystyle{IEEEtran}
\normalem
\bibliography{main}
}


\end{document}